\documentclass[lettersize,journal]{IEEEtran}
\usepackage{amsmath,amsfonts}
\usepackage{algorithmic}
\usepackage{algorithm}
\usepackage{array}
\usepackage[caption=false,font=normalsize,labelfont=sf,textfont=sf]{subfig}
\usepackage{textcomp}
\usepackage{stfloats}
\usepackage{url}
\usepackage{verbatim}
\usepackage{graphicx}
\usepackage{cite}

\usepackage{CJKutf8}
\usepackage{bm}
\usepackage{tabularx}
\usepackage{multirow}
\usepackage{engord}
\usepackage{float}
\usepackage[version=4]{mhchem}
\usepackage[inkscapelatex=false]{svg}
\usepackage{amsfonts} 
\usepackage{pifont}
\newcommand{\cmark}{\ding{51}}%
\newcommand{\xmark}{\ding{55}}%
\newcommand{\ivan}[1]{\textcolor{black}{#1}}
\newcommand{\hr}[1]{\textcolor{black}{#1}}
\newcommand{\hhr}[1]{\textcolor{black}{#1}}
\newcommand{\ivann}[1]{\textcolor{black}{#1}}

\begin{document}

\title{Improved Target-specific Stance Detection on Social Media Platforms by Delving into Conversation Threads}

%
%
%

\author{Yupeng~Li, \IEEEmembership{Member,~IEEE,}
        Haorui~He,
        Shaonan~Wang, \IEEEmembership{Member,~IEEE,}
        Francis~C.M.~Lau, 
        and~Yunya~Song 
\thanks{Yupeng Li is with Department of Interactive Media, Hong Kong Baptist University, Hong Kong (e-mail: ivanypli@gmail.com).}
\thanks{Haorui He is with School of Computer Science, Nanjing University of Posts and Telecommunications, Nanjing, China (e-mail: hehaorui11@gmail.com).}
\thanks{Shaonan Wang is with 
Institute of Automation, Chinese Academy of Sciences, Beijing, China (e-mail: shaonan.wang@nlpr.ia.ac.cn).}
\thanks{Francis C.M. Lau is with Department of Computer Science, The University of Hong Kong, Hong Kong (e-mail: fcmlau@cs.hku.hk).}
\thanks{Yunya Song is with Department of Journalism, Hong Kong Baptist University, Hong Kong (e-mail: yunyasong@hkbu.edu.hk).}
}

\maketitle

\begin{abstract}
Target-specific stance detection on social media, which aims at classifying a textual data instance such as a post or a comment into a stance class of a target issue, has become an emerging opinion mining paradigm of importance. An example application would be to overcome vaccine hesitancy in combating the coronavirus pandemic. However, existing stance detection strategies rely merely on the individual instances which cannot always capture the expressed stance of a given target. In response, we address a new task called conversational stance detection which is to infer the stance towards a given target (e.g., \textit{COVID-19 vaccination}) when given a data instance and its corresponding conversation thread. To tackle the task, we first propose a benchmarking \underline{c}onversational \underline{s}tance \underline{d}etection (CSD) dataset with annotations of stances and the structures of conversation threads among the instances based on six major social media platforms in Hong Kong. To infer the desired stances from both data instances and conversation threads, we propose a model called Branch-BERT that incorporates contextual information in conversation threads. Extensive experiments on our CSD dataset show that our proposed model outperforms all the baseline models that do not make use of contextual information. Specifically, it improves the F1 score by 10.3\% compared with the state-of-the-art method in the SemEval-2016 Task 6 competition. This shows the potential of incorporating rich contextual information on detecting target-specific stances on social media platforms and implies a more practical way to construct future stance detection tasks.
\end{abstract}

\begin{IEEEkeywords}
Target-specific Stance Detection, Opinion Mining, \ivann{Conversation Threads, Social Media Platform}
\end{IEEEkeywords}

\section{Introduction}\label{sec:intro}
\IEEEPARstart{I}{t} is now prevalent for people to express opinions and share their views using social media platforms \cite{phdthesis}, which introduces a new and important channel to discover public stances for commercial and/or research purposes \cite{phdthesis,Tan,TCSS,kuccuk2020stancesurvey,TCSS2}.
For example, the arduous race to increase the vaccine uptake to combat the recent coronavirus pandemic could benefit from detecting the stances towards COVID-19 vaccination in the way where policy makers can formulate suitable policies to overcome vaccine hesitancy. Thus, automatic target-specific stance detection on social media becomes an emerging opinion mining paradigm \cite{stancesurvey2}. 

\begin{figure}[h]
    \centering
    \includegraphics[width=\linewidth]{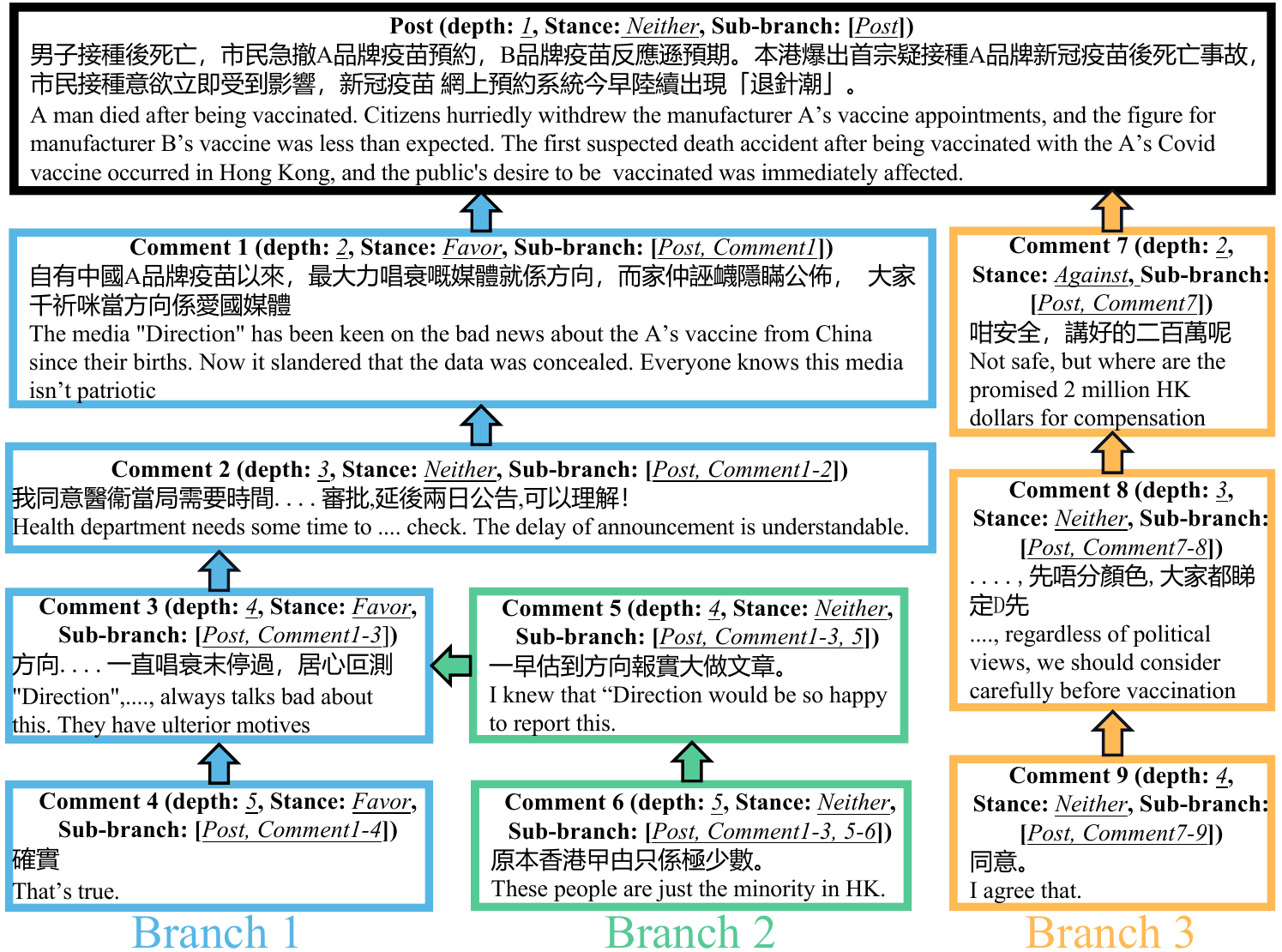}
    \caption{An example of a conversation thread. Due to the space limit, we leave some less critical contents of the instances out. For the complete conversation thread, please see post id 29749172 in our dataset.}
    \label{fig:thread}
\end{figure}

The traditional \textit{target-specific stance detection} \cite{stancesurvey2} on social media platforms classifies a textual data instance, i.e., a post or a comment, into a stance class---\textit{favor}, \textit{against}, or \textit{neither}---based on the stance expressed in the instance towards a given target. Existing methods for target-specific stance detection rely only on an individual's sentences. However, such context-free stance inference is likely unreliable \cite{du2007stance}. For example, Fig.~\ref{fig:thread} illustrates a conversation thread on a Cantonese social platform, \hr{where Comments 4 and 9 show nothing more than a superficial support to the previous comments.} In this case, we can hardly infer the stances of Comments 4 and 9 towards COVID-19 vaccination directly, as the target COVID-19 vaccination was not even mentioned in these comments, \hr{and there is no trigger word about any stance unless the context in previous conversation is also taken into account.} Thus, contextual information in conversation threads is necessary when detecting target-specific stances in comments (such as Comments 1--9 in Fig.~\ref{fig:thread}) in conversation threads. 

To further validate the necessity of contextual information in stance detection on social platforms, we randomly collected 100 conversation threads related to COVID-19 vaccination on six different social platforms. The 100 threads were initiated by 100 posts, accompanied by 1,174 comments in total. We then asked annotators to conduct a two-round annotations. In each round, all the instances were annotated based on the stances of the instances towards the target COVID-19 vaccination. In the first round, only individual instances were provided to the annotators, while in the second round, the individual instances as well as the conversation threads were presented to them.The outcome shows that 48.7\% of instances were mislabeled in the first round but were corrected in the second round, indicating that it is imperative to leverage the contextual information in target-specific stance detection on social media platforms. 

In response, we address a new task called \textit{conversational stance detection}. Given a target entity and a conversation thread on social media platforms, the task is to assign each instance in a conversation thread a stance label, one of \textit{favor}, \textit{against}, and \textit{neither}, which indicates the expressed stance on the given target. This task is challenging to tackle due to the following reasons.
First, there is no existing annotated dataset that addresses our proposed conversational stance detection task. We need to construct a new dataset with annotations of stance and structures of conversation threads among the instances. 
Second, the target of interest may be implicit in the posts or comments. Even if the target is easy to identify, the stance towards the target may be implicit, even when the contextual information is considered. Thus, we need to come up with a smart strategy to incorporate the contextual information in conversation threads to infer the stances of the posts or comments on social media platforms. 

To address the above challenges, we first construct a benchmarking \underline{c}onversational \underline{s}tance \underline{d}etection (CSD) dataset from major social media platforms in Hong Kong,\footnote{We picked Hong Kong because of ease of access to all the mentioned data sources, and also no one has done it for the Cantonese language. Our approach should be applicable to any geo-locations and languages.} which contains 500 Cantonese conversation threads including 500 posts and 5,376 comments, i.e., 5,876 instances. 
In this dataset, the stances of the instances are manually annotated; and the structures of conversation threads among the posts and comments are preserved. Specifically, the timely \textit{COVID-19 vaccination} is the target of interest for data instances and the annotations were performed by 4 annotators. We also employ several techniques to guarantee high-quality annotations and conduct comprehensive evaluations of the baselines for the traditional stance detection task. Furthermore, to extract contextual information in conversation threads on social media platforms for stance inference, we propose a natural language processing model termed Branch-BERT for the target-specific stance detection. 
In this model, each sub-branch in the conversation threads is an input of the BERT model \cite{BERT} in order to generate a sub-branch-specific contextualized text representation for each instance. Then, we employ a TextCNN \cite{TextCNN} model to extract the important $n$-gram features. Lastly, we use a fully-connected neural network to predict the stance in each data instance on the given target. Extensive experiments based on our CSD dataset are conducted to evaluate the performance of our proposed model and the baselines. Though the CSD dataset is in Cantonese, the proposed conversational stance detection task as well as our proposed model are language-independent.

Our contributions are summarized as follows. 
\begin{enumerate}
    \item We propose a task called \textit{conversational stance detection} to detect stance in instances with conversation threads on social media platforms and a benchmarking dataset (i.e., CSD dataset) for the task. To our best knowledge, this is the first dataset for target-specific stance detection which preserves the complete structure of conversation threads on social platforms, leading to a more practical stance detection paradigm on social media platforms.\footnote{The datasets and code for the experiments are available at \url{https://anonymous.4open.science/r/CSD-5A8D/}.} 
    \item We propose a BERT based model called Branch-BERT to address the task. The model provides representations of contextualized text for the comments in the conversations on the social media platforms. With such representations, the model can infer the stances of the posts and the comments on social media platforms. 
    \item Experimental results on our newly constructed CSD dataset show that Branch-BERT outperforms several context-free baselines, including for example the state-of-the-art method \cite{Bertstance} for SemEval-2016 Task 6 (by 10.3\% in F1 score). The results validate that the performance of stance detection methods on social media can be significantly enhanced by incorporating rich contextual information in conversation threads.
    \item To analyze the effectiveness of our proposed model, we evaluate its performance in various cases when the instances have different depths and when the contextual information is partially provided, which may also shed light on how to utilize the contextual information in different cases. In addition, to investigate how the contextual information contributes to the detection outcomes, we conduct a case study that explores what useful contextual information is learned and leveraged in the stance detection task.
\end{enumerate}

The rest of this paper is organized as follows. Sec.~\ref{sec:relate} reviews related studies of stance detection. Sec.~\ref{sec:dataset} and ~\ref{sec:model} introduce our constructed CSD dataset and Branch-BERT model for conversational stance detection task. Sec.~\ref{sec:experiment} demonstrates our experimental results and analysis. Sec.~\ref{sec:conclusion} concludes the paper and discusses important future directions. 

\section{Related Work}\label{sec:relate}
\hhr{There are two typical types of stance detection tasks, \textit{target-specific} and \textit{claim-based} stance detection \cite{stancesurvey2, kuccuk2020stancesurvey}. In this section, we review the two types of stance detection tasks and compare them with our \textit{conversational stance detection} task.}

\subsection{Target-specific Stance Detection}
\hhr{Target-specific stance detection aims to classify a data sample (normally a single sentence) into a stance class in the set $\{\textit{favor}, \textit{against, \textit{neither}}\}$ based on the stance expressed in the text towards a given target (e.g., a person, an organization, a movement, or a policy) independently. 
Recent progress on target-specific stance detection has been promoted by an important dataset, SemEval-2016 Task 6 dataset \cite{mohammad2016dataset}, which consists of tweets (in English) corresponding to five targets during the lead-up stage of the United States 2016 presidential election. 
The dataset contains a training set of 2,814 tweets and a test dataset of 1,249 tweets 
to train supervised stance detection models. There also exist some other similar stance detection datasets in different languages, 
such as English datasets \cite{add1,add2, Coviddata2,data1,data2,Vaccinstance}, Chinese dataset \cite{NLPCC2016overview}, datasets in other languages \cite{taule2017overview,evrard2020overview,Italian,Czech,stance_inter1}, and multi-lingual datasets \cite{vamvas2020multilingual,aaai1,lai2020multilingual}. 
Our conversational stance detection task belongs to the category of target-specific stance detection. We differ from other works in this category in that 
our task leverages the conversation threads on social media platforms instead of single sentences. 
In this work, we propose the first dataset for target-specific stance detection where the conversation threads are preserved, which is also the first stance detection dataset in Cantonese.}\footnote{Cantonese is one of the major Chinese dialects that are widely used in Chinese communities worldwide.}

\hhr{Machine learning algorithms have been proposed to address the target-specific stance detection, e.g., \cite{stance2016mitre,CNNSTANCE,SVM,Tan,BERT}. 
Most of them are based on the SemEval-2016 Task6 dataset \cite{mohammad2016dataset}. 
For example, Zarrella and Marsh \cite{stance2016mitre} used an RNN to predict task-relevant hashtags on a large unlabeled corpus so as to initialize another RNN stance classifier. 
Tackling the same task, another participating model was proposed, which was employed a TextCNN model \cite{CNNSTANCE}. 
Mohammad et al.~\cite{mohammad2017SVM} proposed a powerful baseline that used an SVM model with the word and character level $n$-grams features. This baseline outperformed other models in the competition of SemEval-2016 Task 6. 
In addition, Du et al.~\cite{Tan} used target embeddings and attention mechanisms to incorporate target-specific information to facilitate the stance detection. Li and Caragea \cite{multitask} built a multi-task machine learning model that make use of the sentiment of a tweet to infer its stance. 
Pre-trained language models has shown its dominance over other models in many NLP tasks, especially Bidirectional Encoder Representations from Transformers (BERT) \cite{BERT}. 
To tackle the target-specific stance detection, Ghosh et al.~\cite{Bertstance} used a fine-tuned BERT model 
that achieved the state-of-the-art performance of an F1 score of 75.6\% on SemEval-2016 Task 6 dataset. 
However, these methods all 
ignore the helpful contextual information in conversation threads. 
Different from the above works, we propose a Branch-BERT model that takes the advantage of the contextual information which can 
infer the stances of a certain instance (e.g., a reply to a tweet post) implicitly embedded in the conversation threads (see our example in Fig.~\ref{fig:thread}).} 

\subsection{Claim-based Stance Detection} 
\hhr{Different from the above target-specific stance detection, claim-based stance detection aims to detect the stance in \ivann{a comment} to a claim in a piece of news toward whether the comment confirms the claim or challenges its validity, with applications to rumour validation and fake news detection. The prediction classes are usually in \{\textit{Supporting}, \textit{Denying}, \textit{Querying}, \textit{Commenting}\} \cite{rumoureval2017,zubiaga2018rumoursurvey,RumourEval2019}.}

\hhr{This task is regarded as a \emph{sentence-pair classification} task, which is different from the aforementioned target-specific stance detection whose objective is, as addressed above, to detect the stance in an instance, either a post or a comment, towards a specific target entity, e.g., \textit{COVID-19 vaccination} and \textit{Donald Trump}. 
Note that some claim-based stance detection methods utilized the tree-structured conversations that contain the target comment whose stance needs to be inferred \cite{Branchlstm,fajcik2019but,zubiaga2018discourse}. However, they were proposed for the sentence-pair classification task (i.e., the stance of a comment toward the claim in the post), based on the nature of the claim-based stance detection task. 
Different from claim-based stance detection, our conversational stance detection task belongs to the category of target-specific stance detection. 
In addition, we 
use the post and other comments in the conversational threads as auxiliary information to classify the stance of each instance (a post or a comment) towards the given target. Claim-based stance detection methods can hardly be leveraged to address our task.}

\hr{Table~\ref{tab:concept} summarises the differences between our task and the two types of traditional stance detection.}
\begin{table}[h]
\centering
\caption{Comparison of different stance detection tasks.}
\begin{tabular}{|c|cc|c|}
\hline
\textbf{Type} & \multicolumn{2}{c|}{\textbf{Target-specific}} & \textbf{Claim-based} \\ \hline
\textbf{Classif.~Task} & \multicolumn{2}{c|}{\begin{tabular}[c]{@{}c@{}}Sentence \\ classification\end{tabular}} & 
\begin{tabular}[c]{@{}c@{}}Sentence-pair\\ classification\end{tabular} \\ \hline
\textbf{Objective} & \multicolumn{2}{c|}{\begin{tabular}[c]{@{}c@{}}Detect the stance of a \\post or a comment \\toward \ivann{an \underline{\emph{entity}}}\end{tabular}} & \begin{tabular}[c]{@{}c@{}}Detect the stance of \\a comment toward the \\\underline{\emph{claim}} in the post\end{tabular} \\ \hline
\textbf{\begin{tabular}[c]{@{}c@{}}Example of \\Entity or Claim\end{tabular}} & \multicolumn{2}{c|}{\begin{tabular}[c]{@{}c@{}}COVID-19  \\Vaccination\end{tabular}} & 
\begin{tabular}[c]{@{}c@{}}``We understand there are \\two gunmen and up  to \\a dozen hostages inside \\ the cafe under siege at \\Sydney.. ISIS flags \\remain on display..'' \end{tabular} \\ \hline
\textbf{Label} & \multicolumn{2}{c|}{\begin{tabular}[c]{@{}c@{}}Favor, Against,\\ Neither\end{tabular}} & \begin{tabular}[c]{@{}c@{}}Supporting, Denying,\\ Querying, Commenting\end{tabular} \\ \hline
\textbf{\begin{tabular}[c]{@{}c@{}}Use Convo.\\Threads\end{tabular}} & \multicolumn{1}{c|}{\xmark} & \cmark & \cmark \\ \hline
\textbf{Work} & \multicolumn{1}{c|}{\begin{tabular}[c]{@{}c@{}}e.g.,
\cite{mohammad2016dataset,NLPCC2016overview},\\\cite{SVM,Tan,BERT}\end{tabular}} & \begin{tabular}[c]{@{}c@{}}Our \\work\end{tabular} & \begin{tabular}[c]{@{}c@{}}e.g.,
\cite{rumoureval2017,zubiaga2018rumoursurvey,RumourEval2019,Branchlstm,fajcik2019but,zubiaga2018discourse} \end{tabular} \\ \hline
\end{tabular}
\label{tab:concept}
\end{table}

\section{Conversational Stance Detection Dataset}\label{sec:dataset}
\begin{figure}[h]
    \includegraphics[width=\linewidth]{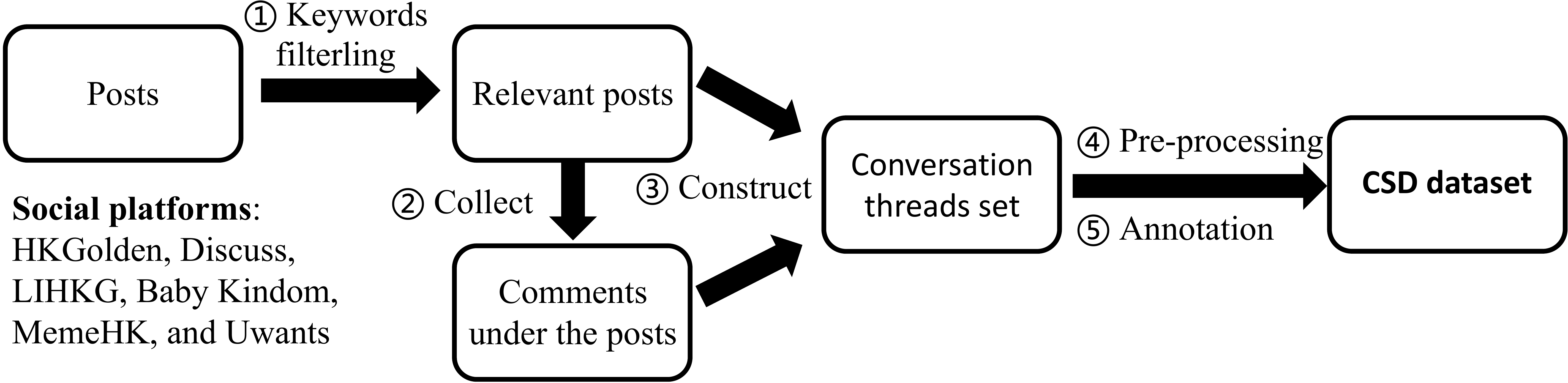}
    \caption{The process of CSD dataset construction.}
    \centering
    \label{fig:datasetpip}
\end{figure}

To our knowledge, there is no existing dataset for target-specific stance detection that captures \ivan{the contextual information in conversation threads on social media platforms}.
Thus, to tackle the conversational stance detection task, i.e., to design machine learning based solutions, we construct a new dataset called \emph{\underline{c}onversational \underline{s}tance \underline{d}etection} (CSD) dataset. In this dataset, we selected the prevalent \emph{COVID-19 vaccination} as our target of interest. Fig.~\ref{fig:datasetpip} demonstrates the five-step procedure to construct the dataset.

\subsection{Dataset Overview}
The CSD dataset consists of 500 conversation threads (including 500 posts and 5376 comments) from six major social media platforms in Hong Kong. Table~\ref{tab:Dataset_statistic} illustrates some statistics of our CSD dataset, including the average Cantonese character count, and the number of instances with different depths. Note that the size of our dataset is larger than prevalent datasets for target-specific stance detection, i.e., the SemEval-2016 Task 6 dataset \cite{mohammad2016dataset} and the NLPCC-ICCPOL-2016-Shared-Task4 dataset \cite{NLPCC2016overview}.

A conversation thread refers to a conversation on a social platform that is initiated by a post and contains zero, one, or more comments. Fig.~\ref{fig:thread} provides an example of a conversation thread. We use the tree structure to model conversation threads, where each instance is represented as a node in a tree with the post as the root node. For example, in Fig.~\ref{fig:thread}, Comment 1 is the parent node of Comment 2 since Comment 2 replies to Comment 1. Such inter-instance relationships are recorded in our dataset which can be used to reconstruct the original structure of each conversation thread. The unique path from the root node to each leaf node is called a branch of the tree. The path from the root node to each node within a branch is called a sub-branch. The depth of a node is the length of its sub-branch. For example, the depths of the Post, Comment 1, and Comment 2 are 1, 2, and 3, respectively.
\begin{table}[h]
\caption{Statistics of the CSD dataset.}
    \begin{tabular}{cccc}
        \hline
        \textbf{Instance} & \textbf{Avg.~char count} & \textbf{Depth} & \textbf{Number} \\
        \hline
        Post & 223.18 & 1 & 500 (8.5\%) \\
        \hline
        \multirow{4}{*}{Comment} & 20.19 & 2 & 4205 (71.6\%) \\
         & 19.10 & 3 & 798 (13.6\%)\\
         & 21.13 & 4 & 238 (4.1\%)\\
         & 17.01 & $\geq5$ & 135 (2.2\%)\\
         \hline
    \end{tabular}
\label{tab:Dataset_statistic}
\centering
\end{table}

\subsection{Data Collection (Step 1 - Step 4)}
 As shown in Fig.~\ref{fig:datasetpip}, we first collected posts from six major social media platforms in Hong Kong, HKGolden, Discuss, LIHKG, Baby Kindom, MemeHK, and Uwants.\footnote{HKGolden: \url{https://m.hkgolden.com/}; Discuss: \url{https://www.discuss.com.hk/}; LIHKG: \url{https://lihkg.com/}; BabyKindom: \url{https://www.baby-kingdom.com/}; MemeHK: \url{https://forum.memehk.com}; Uwants: \url{https://www.uwants.com/}} Using the keywords listed in Table~\ref{tab:keywords}, we filtered out (\underline{Step 1}) those relevant posts, which were posted between 23 December 2020\footnote{The day when the Chief Executive of Hong Kong announced for the first time that the Hong Kong government had purchased 22.5 million doses of COVID-19 vaccines and promulgated relevant regulations.} and 15 August 2021. Afterward, a manual selection was conducted to remove unrelated posts. In \underline{Step 2}, we randomly sampled 500 posts from the target-related post set and crawled all comments under these posts as the comment set. Inter-instance relationships in these posts and comments are recorded in \underline{Step 3} to construct a tree-structured conversation thread set. Apart from comments that express opinions towards our target, there are several comments in which the target of opinion is not our stance target in the comment set. These comments should be classified as ``neither''. Including these instances further complicates our CSD task.

\begin{table}[h]
    \centering
    \caption{Keywords to filter relevant posts.}
    \begin{tabular}{c}
        \hline
        \begin{CJK*}{UTF8}{gbsn}疫苗, 免疫, 科兴, 复必泰, 北京生物, 武汉生物, \end{CJK*} \\ \begin{CJK*}{UTF8}{gbsn}辉瑞,莫德纳, 克尔来福, 复星, 阿斯利康, 不良反应, \end{CJK*} \\
        \begin{CJK*}{UTF8}{gbsn} 蛋白, 谷针, 一针, 两针, 接种, 打针, 灭活, 副作用, \end{CJK*}\\ 
        MRNA, VACCIN, VAXX,IMMUNIZATION,PFIZER,\\ 
        IMMUNO, SUPPRESSED, MODERNA, IMMUNE,\\ SINOVAC, CORONAVAC, COMIRNATY, BIONTECH,\\
        ASTRAZENECA,IMMUNE, INOCULATION\\
        \hline
    \end{tabular}
    \label{tab:keywords}
\end{table}
 
Since messages on social media platforms are frequently written in an informal style, pre-processing of online textual data is needed for stance analysis \cite{Preprocess1,Preprocess2,preprocess3}. 
Specifically in \underline{Step 4}, we first clean the data by removing redundancies such as the HTML tags, punctuation, white spaces, links, etc. Emojis were replaced with the corresponding words as well. Then, we leveraged the OpenCC library to convert simplified Chinese characters to Hong Kong standard traditional Chinese characters for text standardization.\footnote{\url{https://github.com/BYVoid/OpenCC}} 
In addition, duplicated instances were discarded, as well as the null ones. 

\subsection{Data Annotation (Step 5)}
To annotate the dataset, we leveraged the definition and categories of stance in the SemEval-2016 Task 6 \cite{Semeval2016}. Each instance should be assigned a label from the set $\{\text{favor}, \text{against, \text{neither}}\}$. Mimicking how people actually use social platforms, the conversation threads were presented to four independent annotators, who are all native Cantonese speakers. We trained the annotators for a few rounds of pilot annotations before they started to annotate the dataset. For reliability, each instance was annotated by at least three annotators. In the case of disagreement, the final labels of the instances were decided by in-depth group discussion.

Table~\ref{tab:depth_stance} demonstrates the distribution of instances with different depths and stance labels. As shown, 36.9\% of the instances express an opposing stance towards the target ``COVID-19 vaccination'', while only 14.2\% of instances are in favor of it. Nearly half of the instances do not have a biased attitude towards the target of interest. Moreover, there is an obvious trend that as the depth increases, the proportion of \textit{neither} instances increases (from 24\% to more than 58\%), while the proportion of \textit{favor} and \textit{against} instances decreases (from 30.6\% to around 8.0\% and from 45.4\% to around 31.0\%, respectively). This means that instances with deeper depths are more unbalanced in their stance labels, and, thus, their stances may be more difficult to detect.

\begin{table}[h]
\caption{The distribution of instances with different depths and stance labels.}
    \begin{tabular}{ccccc}
    \hline
    \textbf{Depth} & \textbf{Favor} & \textbf{Against} & \textbf{Neither}  \\
    \hline
    1 & 153 (30.6\%) & 227 (45.4\%) & 120 (24.0\%)  \\
    2 & 591 (14.0\%) & 1575 (37.5\%) & 2039 (48.5\%)  \\
    3 & 62 (7.8\%) & 249 (31.2\%) & 487 (61.0\%)\\
    4 & 17 (7.1\%) & 72 (30.3\%) & 149 (62.6\%)  \\
    $\geq$5 & 12 (8.9\%) & 44 (32.6\%) & 79 (58.5\%) \\
    \hline
    Total & 835 (14.2\%) & 2167 (36.9\%) & 2874 (48.9\%) \\
    \hline
    \end{tabular}
\label{tab:depth_stance}
\centering
\end{table}
\section{The Branch-BERT Model}\label{sec:model}
\begin{figure*}[t]
    \centering
    \includegraphics[width=\linewidth]{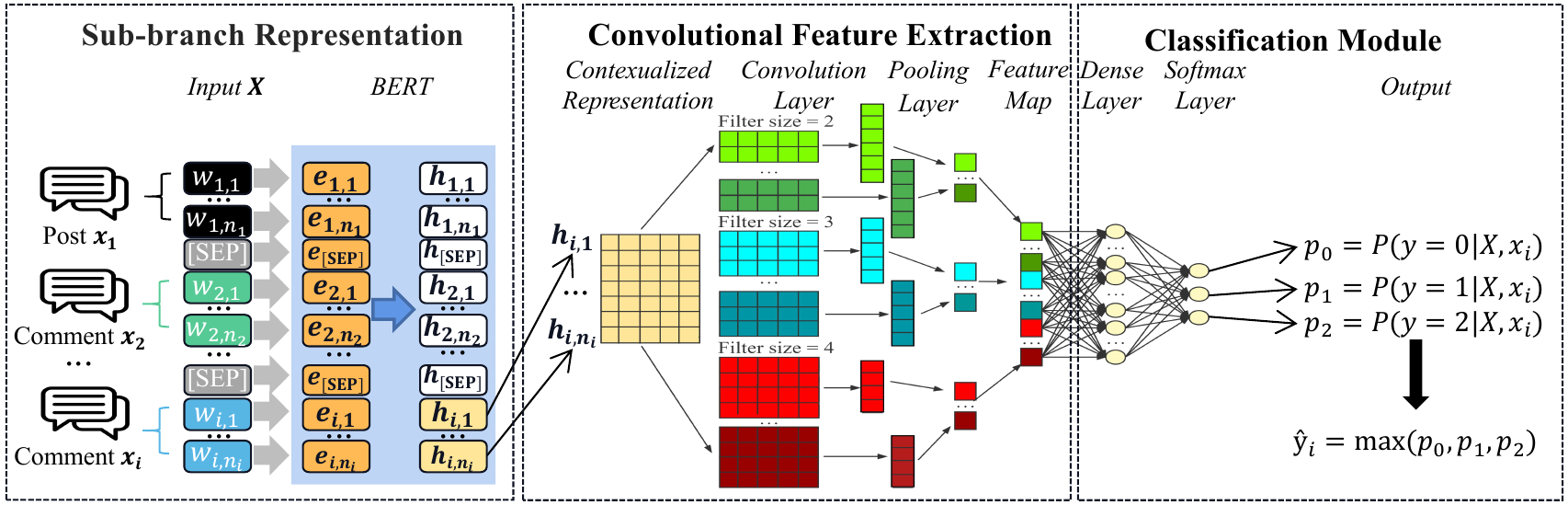}
    \caption{Architecture of Branch-BERT, whose input is the sub-branch $\bm{B}$ of instance $\bm{x}_{i}$, output is its predicted label $\hat{y}_i$.}
    \label{fig:model}
\end{figure*}
As discussed, the performance of stance detection can be potentially improved by considering the contextual information in conversation threads. Thus, we propose a model termed Branch-BERT for our conversational stance detection task, which can extract contextualized semantics in conversation threads. As shown in Fig.~\ref{fig:model}, our proposed model contains three modules: (1) \emph{Sub-branch Representation Module} employing BERT \cite{BERT} as the backbone to generate a contextualized representation of each token in the text of an input sub-branch;\footnote{Note that the model requires no label information from any other instances.} (2) \emph{Convolutional Feature Extraction Module} utilizing a TextCNN \cite{TextCNN} to extract important $n$-grams features from output of Sub-branch Representation Module; (3) \emph{Classification Module} that determines the final stance labels. The details of these three modules are presented as follows.

\subsection{Sub-branch Representation Module} \label{subsec:GAP}
Context-independent word embeddings like Word2Vec \cite{Word2vec} or GloVe \cite{glove} for each token in a text cannot effectively give deep contextualized representations \cite{ELMOS}. Therefore, we use BERT to generate deep contextualized representations for sub-branches in the conversation threads.

We denote a sub-branch 
$\bm{B} = [\bm{x}_{1},\cdots,\bm{x}_{i}]$, where 
$\bm{x}_{u}= [w_{u,1},\cdots,w_{u,n_{u}}]$ 
($\forall u=1,\cdots,i$) is an instance representing a post or a comment. There are $n_u$ tokens in $\bm{x}_{u}$, each of which $w_{u,v}$ ($\forall v = 1,\cdots,n_u$) denotes the $v$-th token in $\bm{x}_{u}$. Note that $\bm{x}_{1}$ and $\bm{x}_{u} (\forall u=2,\cdots,i)$ denote the post and the comment with depth $u$ in this sub-branch, respectively.

For an instance $\bm{x}_{i}$, to extract the contextual information carried by ancestor nodes (instances)\footnote{\hr{In some cases, we need to infer an instance's stance immediately when it is posted. Thus, we consider the contextual information in the ancestor nodes to generally capture such practical need.}} in the corresponding sub-branch $\bm{B}$, we concatenate all instances in $\bm{B}$ into a token sequence $\bm{X}$ in which every two consecutive instances are separated with a special token [SEP].
Let $\oplus$ be the concatenation operation. Thus, we have $$\bm{X} = \bm{x}_{1} \oplus [\text{SEP}] \oplus \bm{x}_{2} \oplus \cdots \oplus [\text{SEP}] \oplus \bm{x}_{i}.$$ Then the length of $\bm{X}$ is $$l = \sum_{k=1}^i n_{k}+i-1.$$ Before input to BERT, $\bm{X}$ needs to be pre-processed by a BERT tokenizer, whose input length is upper bounded by 512 tokens. However, $l$ is likely to exceed the length limit because we concatenated multiple instances in a sub-branch. As shown in Table~\ref{tab:Dataset_statistic}, the average number of characters of posts is about 10 times more than that of comments. This may be due to the reason that a post is more likely to be a news piece or long statement. Therefore, if $l > 512$, $\bm{x}_{1}$ would be replaced by its abstract which is automatically generated \cite{auto-abstracts}.

Then, we use a BERT tokenizer to convert $\bm{X}$ to BERT's input embeddings $\bm{E}$, which is the sum of token embeddings $\bm{E}_{t}$, segment embeddings $\bm{E}_{s}$, and positional embeddings $\bm{E}_{p}$: $$\bm{E} = \bm{E}_{t} + \bm{E}_{s} + \bm{E}_{p} = [\bm{e}_{1,1},\cdots,\bm{e}_{1,n_{1}},\cdots,\bm{e}_{i,1},\cdots,\bm{e}_{i,n_{i}}],$$ where $\bm{e}_{i,j}$ ($j \in [1,n_{i}]$) denotes the input embedding corresponding to token $w_{i,j}$. Given the input $\bm{E}$, the final hidden state $\bm{H} \in \mathbb{R}^{l \times h}$ (where $h$ is the hidden size of BERT) output by BERT is 
$$\bm{H} = [\bm{h}_{1,1},\cdots,\bm{h}_{1,n_{1}},\cdots,\bm{h}_{i,1},\cdots,\bm{h}_{i,n_{i}}],$$ where $\bm{h}_{i,j}$ denotes the token vector of  $\bm{w}_{i,j}$. We use the sub-sequence $[\bm{h}_{i,1},\cdots,\bm{h}_{i,n_{i}}]$ to be the contextualized text representation of $\bm{x}_{i}$. Because $n_{i}$ is a varying value, we conduct zero-padding or cutting to ensure every sequence is in same size $d$. Then, the final contextualized text representation $\bm{H}_{\bm{x}_{i}} \in \mathbb{R}^{d \times h}$ of $\bm{x}_{i}$: $$\bm{H}_{\bm{x}_{i}} = [\bm{h}_{i,1},\cdots,\bm{h}_{i,d}],$$ which is the input of the next Convolution Feature Extraction Module or is sent to the last Classification module after global average pooling. 

\subsection{Convolution Feature Extraction Module}
Our BERT tokenizer when splitting Cantonese text into characters may incur, a loss in semantic information carried by larger text units like words and phrases \cite{WordBERT, ZEN}. To reduce the information loss, we employ a TextCNN model to be our $n$-gram feature extractor.

A convolution operation in the module involves a filter $\bm{w} \in \mathbb{R}^{h \times k}$, which is applied to a window of $k$ token vectors to generate a feature. For example, the feature $f_{j}$ is generated from the $j$-th window of token vectors $\bm{H}_{\bm{x}_{i},j} = [\bm{h}_{i,j},\cdots,\bm{h}_{i,j+k-1}]$ in $\bm{H}_{\bm{x}_{i}}$ 
by $$ f_{j} = \text{ReLU}(\bm{w}\cdot\bm{H}_{\bm{x}_{i},j}+b_{1}),$$ where $\text{ReLU}$ is a non-linear activation function and $b_{1}$ is a bias vector. After applying $\bm{w}$ to every possible window of token vectors in $\bm{H}_{\bm{x}_{i}}$, the filter $\bm{w}$ produces a feature map $$\bm{F} = [f_{1},\cdots, f_{n_{i}-k+1}].$$
We then apply a max-pooling operation \cite{maxpooling} over the feature map $\bm{F}$ to take the maximum value $$\hat{f}= \max(\bm{F})$$ as the extracted important $n$-gram feature by $\bm{w}$. Multiple filters with varying window sizes are employed in this module to obtain multiple features. The generated features are denoted by $$\bm{z} = [\hat{f}_{1},\cdots,\hat{f}_{p}],$$
where $p$ is the number of filters.

\subsection{Classification Module}
Given the generated features $\bm{z}$ by the Convolution Feature Extraction Module, 
we first adopt a dropout layer to avoid over-fitting: 
$$\bm{\hat{z}} = \text{Dropout}(\bm{z}),$$
whose output $\bm{\hat{z}}$ is then passed to a fully connected layer and a softmax layer to compute the probability distribution $\bm{p}$ over the three potential labels of stances:
$$\bm{p} = \text{softmax}(\bm{W}\cdot\bm{\hat{z}}+b_{2}),$$
where $\bm{W} \in \mathbb{R}^{3 \times p}$ is the weight matrix in the fully connected layer, and $b_{2}$ is a bias term. Finally, the class having the maximum value is chosen as the predicted label $\hat{y}$ for the input instance (post or comment) $\bm{x}_{i}$: $$\hat{y} = \max{(\bm{p})}.$$

\section{Experiments}\label{sec:experiment}
We conduct extensive experiments on the proposed CSD dataset in this section to validate the effectiveness of contextual information in conversation threads. In this section, we present detailed experimental settings, including carefully chosen baselines and the parameters used (Sec.~\ref{sec:baseline}), followed by the result analysis (Sec.~\ref{sec:exp_result}-\ref{sec:case}).

\subsection{Experiment Settings}\label{sec:baseline}
As stated, existing works on target-specific stance detection leverage only textual contents in each instance without considering contextual information. Four such baselines are carefully selected for performance comparison. The details such as their implementations on the CSD dataset are illustrated as follows: 
\begin{itemize}
    \item \textbf{SVM+$\mathbf{n}$-grams} \cite{mohammad2017SVM}: a SVM classifier uses word-level and character-level $\mathbf{n}$-grams features. The model outperformed all other solutions for the sub-task A in the SemEval-2016 Task 6 competition \cite{Semeval2016}. To implement the method on the CSD dataset, we leveraged the Pycantonese \cite{pycantonese} library for Cantonese word segmentation.
    \item \textbf{TextCNN} \cite{CNNSTANCE}: the model ranked the $2^{nd}$ and the $1^{st}$ respectively on sub-task A and sub-task B in the SemEval-2016 Task 6 competition among all solutions. To adapt the baseline on the CSD dataset, we used the Word2Vec \cite{Word2vec} to initialize all word vectors. The word embedding vectors are pre-trained on unlabelled Chinese Wikipedia corpora.
    \item \textbf{Tan} \cite{Tan}: the model uses an attention mechanism to learn the correlation between the texts and the given target. In our re-implementation, the given target is set as \begin{CJK*}{UTF8}{gbsn} ``接种新冠疫苗''\end{CJK*} (``COVID-19 vaccination''). The baseline was initialized by the same word vectors as the previous baseline.
    \item \textbf{BERT} \cite{Bertstance}: A fine-tuned BERT model for stance detection, which yielded the state-of-the-art result on the SemEval-2016 Task 6 sub-task A. Because there is no Cantonese BERT model publicly available, we first build a Cantonese corpus based on the collected Cantonese text from online social platforms, the Hong Kong Cantonese corpus\footnote{http://compling.hss.ntu.edu.sg/hkcancor}, and the Hong Kong mid-twentieth century Cantonese corpus\footnote{http://corpus.ied.edu.hk/hkcc}, and then use the corpus to pre-train a BERT base model. 
\end{itemize}

\begin{table}[h]
\centering
\caption{Model-wise hyper-parameter settings adopted in our (re-)implementations of the baselines and our model.}
\begin{tabular}{ll}
\hline
\textbf{Model} & \textbf{Hyper-parameters} \\
\hline
SVM+$\mathbf{n}$-grams & 
\begin{tabular}[c]{@{}l@{}}word-level $\mathbf{n}$-grams: 1,2,3;\\ character-level $\mathbf{n}$-grams: 2,3,4,5.\end{tabular} \\
\hline
TextCNN & \begin{tabular}[c]{@{}l@{}}padding size: 64; \\learning rate: 5e-4; \\ filter sizes: (2, 3, 4);\\ number of fliters: 32.\end{tabular} \\
\hline
Tan & \begin{tabular}[c]{@{}l@{}}padding size: 64;\\ number of LSTM layers: 2;\\
LSTM hidden size: 256;\\ learning rate: 5e-4.\end{tabular}\\
\hline
BERT & \begin{tabular}[c]{@{}l@{}}padding size: 64\\; learning rate: 1e-5.\end{tabular}\\
\hline
Branch-BERT & \begin{tabular}[c]{@{}l@{}} BERT learning rate: 1e-5;\\ CNN learning rate: 1e-4; \\ filter sizes: (2,3,4);\\number of filters: 32; \\zero-padding size: 64.\end{tabular}\\
\hline
\end{tabular}
\label{tab:parameter}
\end{table}

Table~\ref{tab:parameter} demonstrates model-wise hyper-parameter settings adopted in our implementations of our model and the baselines. Please refer to the respective papers for those unmentioned parameters. We compare the performance of the baselines and our proposed Branch-BERT model. Specifically, we use the Adam optimizer to train all the models. An early stopping strategy is adopted in the model training. 
The batch size is set as 16. We stop training the models if no improvement is identified after 500 consecutive batches which is enough to reach the convergence. Similar to other text classification studies \cite{mohammad2017SVM, NLPCC2016overview, Bertstance}, we evaluate the performance of models in the macro-averaged F1 score. The ratio of the train set and test set is 8:2. All experiments were repeated ten times on Google Colaboratory with a single Tesla P100-PCIE-16GB GPU. We took the average of the experimental results for analysis (see Sec.~\ref{sec:exp_result}).
\begin{table*}[h]
\caption{The overall performance of the baselines and Branch-BERT model with different configurations, and their performance on instances with different depths. The evaluation metric is macro-averaged F1 score. The two numbers in each entry are the mean and the standard deviation of the F1 scores. }
\begin{tabular}{lllllll}
\hline
\textbf{Models} & \textbf{Overall}&\textbf{Depth=1} & \textbf{Depth=2} & \textbf{Depth=3} & \textbf{Depth=4} & \textbf{Depth$\geq$5} \\
\hline
SVM+$n$-grams & 0.581$\pm$0.015& 0.601$\pm$0.067 & 0.575$\pm$0.018 & 0.500$\pm$0.051 & 0.516$\pm$0.100 & 0.473$\pm$0.113 \\
TextCNN & 0.539$\pm$0.018 &0.533$\pm$0.062 & 0.526$\pm$0.029 & 0.457$\pm$0.038 & 0.469$\pm$0.096 & 0.428$\pm$0.101 \\
Tan & 0.531$\pm$0.009 &0.541$\pm$0.052 & 0.521$\pm$0.012 & 0.434$\pm$0.037 & 0.442$\pm$0.106 & 0.459$\pm$0.131 \\
BERT & 0.608$\pm$0.021 
&0.586$\pm$0.065 
& 0.609$\pm$0.028 
& 0.485$\pm$0.053 & 0.484$\pm$0.118 & 0.450$\pm$0.084\\
\hline
\textbf{Our Configurations} & & & & & & \multicolumn{1}{c}{} \\
\hline
Branch-BERT (w/o SR) & 0.613$\pm$0.022 & 0.607$\pm$0.065 & 
0.615$\pm$0.032 & 
0.490$\pm$0.070 & 0.497$\pm$0.146 & 0.446$\pm$0.064 \\
Branch-BERT (w/o CFE) & 0.690$\pm$0.012 & 0.638$\pm$0.053 & 0.692$\pm$0.020 & 0.615$\pm$0.068 & \textbf{0.601$\pm$0.118} & 0.500$\pm$0.099 \\
Branch-BERT & \textbf{0.711$\pm$0.018} &\textbf{0.658$\pm$0.045} & \textbf{0.715$\pm$0.024} & \textbf{0.639$\pm$0.051} & 0.570$\pm$0.101 & \textbf{0.588$\pm$0.166} \\
\hline
\end{tabular}
\label{tab:depth_results}
\centering
\end{table*}
\subsection{Overall Performance Evaluation}\label{sec:exp_result}
As shown in Table~\ref{tab:depth_results}, among all the baselines, the fine-tuned BERT model performs the best with an F1 score of 60.8\%. This result is consistent with the evaluation results on the SemEval-2016 Task 6 that the fine-tuned BERT \cite{Bertstance} is the best model for target-specific stance detection. In comparison, our proposed Branch-BERT model can achieve an F1 score of 71.1\%, which outperforms substantially all the baselines. Specifically, Branch-BERT outperforms the BERT model by 10.3\%. Our results validate the necessity of contextual information in conversation threads, which is ignored by all the baselines for the task. 

\subsection{Ablation Study}\label{sec:ablation}
To further study the specific contributions of the Branch-BERT model to our task, we conducted an ablation study to evaluate other two configurations of our proposed model. Compared to Branch-BERT, the first configuration discards the Sub-branch Representation (SR) Module in Branch-BERT. This configuration ignores the contextual information in conversation threads when classifying each instance into a stance. We refer to the configuration as Branch-BERT (w/o SR). The second configuration differs from Branch-BERT in that it discards the Convolution Feature Extraction (CFE) Module. That is, the output of the Branch Representation Module is directly sent to the Classification Module after the global average pooling. We refer to the configuration as Branch-BERT (w/o CFE).

We can observe from Table~\ref{tab:depth_results} that both configurations do not perform as well as Branch-BERT. This ablation study demonstrates that both the SR and CFE modules contribute to the target-specific stance detection. Comparing these two modules, Branch-BERT outperforms the Branch-BERT (w/o CFE) configuration (resp.~Branch-BERT (w/o SR) configuration) by 2.1\% (resp.~9.8\%). The result indicates that the contextualized text representation of sub-branches in conversation threads generated by the SR module is the key to the considerable improvement of our model. The CFE module also brings benefit to our model. This may due to that the CFE module effectively extracts important $n$-gram features to alleviate the loss in semantic information carried by larger text units like words and phrases that was brought by Cantonese BERT tokenizer \cite{ZEN}.  

\subsection{Impact of Depth of the Instance}\label{sec:depth}
\begin{figure}[h]
    \centering
    \includegraphics[width=\linewidth]{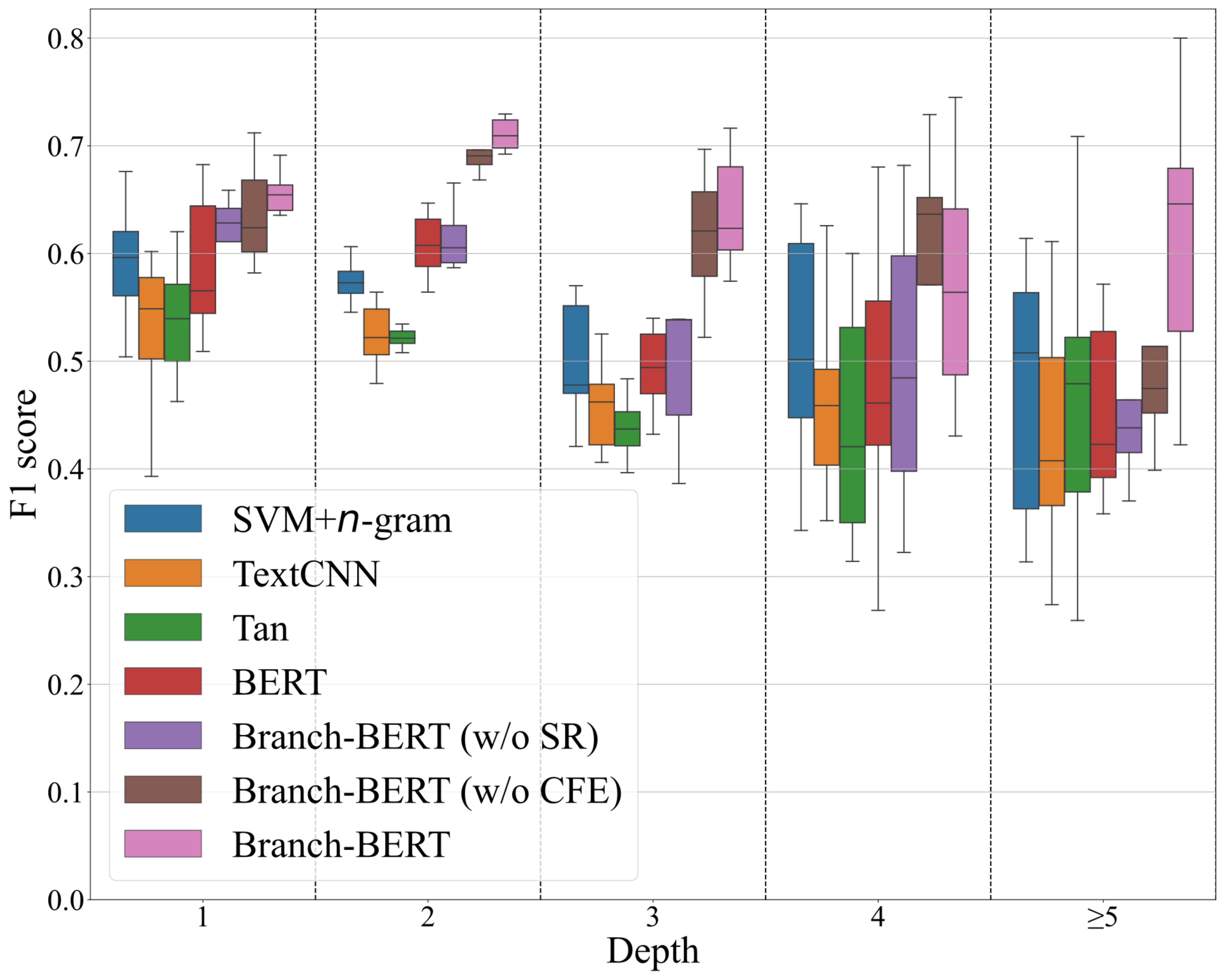}
    \caption{The performance of the baselines and Branch-BERT on instances with different depths.}
    \label{fig:depth_performances}
\end{figure}

As illustrated in Table~\ref{tab:Dataset_statistic}, the instances in the CSD dataset can be categorized into five classes, based on their depths, i.e., instances with a depth of 1, 2, 3, 4 or $\geq$5. In this part, we conduct experiments to analyze the impact of the depth on the performance of the baselines and Branch-BERT. Fig.~\ref{fig:depth_performances} and Table \ref{tab:depth_results} summarize the experimental results. Our results show that Branch-BERT outperforms all the methods that do not consider the contextual information in conversation threads, i.e., the baselines with any depths. Specifically, the performance improvement of our model compared with the baselines is more significantly in the cases with a depth of 2 or 3 that those with a depth of 4+.

\hr{When the depth is 1, Branch-BERT outperforms the best context-free model (i.e., Branch-BERT (w/o SR) by around 5.1\%. Although they use the same inputs (i.e., the posts) in the inference, Branch-BERT, when training with the comments (i.e., instances with depths $\geq 2$), may learn the information from the posts in the same sub-branches that is useful for the stance detection task. This process may in turn improve the performance of inferring the stance of the posts themselves. 
This finding implies that learning from the context (i.e., the sub-branches) can be beneficial to detecting stances of the instances in the context. 
When the depth is 2 or 3, the advantages of our model and Branch-BERT (w/o CFE) become prominent. Specifically, the Branch-BERT model outperforms Branch-BERT (w/o SR) in F1 score by 10\% and 14.9\% when the depth is 2 and 3, respectively.} As demonstrated in Table~\ref{tab:Dataset_statistic}, instances with depth of 2 or 3 account for 85.2\% of the instances on social media platforms. Hence, the performance improvement of Branch-BERT in these two cases compared with context-free models validates the effectiveness of the contextual information in conversation threads in target-specific stance detection. 
When the depth $\geq 4$, the performance variation of any method is more significant than that in other cases. This may be because the contextual information that instances with higher depths require may be generated by more than one user on the social media platforms, which further complicates the model learning, in particular, when extracting the semantic information.  

\subsection{Using Partial Contextual Information}
In this part, we investigate stance detection using partial contextual information (instead of full information) in the conversation. Specifically, we consider a variant of our CSD task which leverages a \emph{slice} of the sub-branch of a data instance in both the training and the inference stages. 
Recall that $\bm{B} = [\bm{x}_{1},\cdots,\bm{x}_{i}]$ is the sub-branch corresponding to instance $\bm{x}_{i}$. 
We define the partial sub-branch $\bm{B}_{k}$ for instance $i$ as follows. 
 $$
\bm{B}_{k} =
\begin{cases}
[\bm{x}_{1},\cdots,\bm{x}_{i}],  & \text{if $i \leq k$} \\
[\bm{x}_{i-k},\cdots,\bm{x}_{i}], & \text{if $i > k$}
\end{cases},
$$
where  $k$ is the maximum possible number of ancestor nodes (instances) that can be utilized in the context. For example, in branch 1 in Fig.~\ref{fig:thread}, $\bm{B}_{2}$ for Comment 1 is $[\text{Post }, \text{Comment } 1]$; $\bm{B}_{2}$ for Comment 4 is $[\text{Comment } 2, \text{Comment } 3, \text{Comment } 4]$. 

We evaluate the performance of Branch-BERT using four different inputs, $\bm{B}_{0}$, $\bm{B}_{1}$, $\bm{B}_{2}$ and $\bm{B}$, which correspond respectively to the different levels of the available contextual information. Note that the models with the input of $\bm{B}_{0}$ (using only the instance itself) and full sub-branches $\bm{B}$ are equivalent to the Branch-BERT (w/o SR) and the Branch-BERT, respectively. As shown in Table~\ref{tab:part}, the model trained with $\bm{B}$ outperforms the models using the partial contextual information in F1 score. Specifically, the performance of our model improves as $k$ increases.  
The result indicates that the stance detection benefits more from more available contextual information. Noticeably, the model with the input of $\bm{B}_{1}$ achieves an F1 score of 0.698, which performs a bit less than the models with more contextual information (i.e., $\bm{B}_{2}$ and $\bm{B}$) but significantly outperforms the one using the instance itself (i.e., $\bm{B}_{0}$). The above qualitative evaluations show a trade-off between the performance of our stance detection model and available contextual information, which sheds light on using partial contextual information in certain scenarios, 
in particular, when the available computing resources for model training and/or inference are constrained.

\begin{table}[h]
\centering
\caption{The performance of Branch-BERT with different extents of the available (partial) contextual information.}
\begin{tabular}{lc}
\hline
\textbf{Input} & \textbf{F1 Score} \\
\hline
$\bm{B}_{0}$ (using only instance itself) & 0.613$\pm$0.022\\
$\bm{B}_{1}$ (partial contextual information)& 0.698$\pm$0.014 \\
$\bm{B}_{2}$ (partial contextual information)& 0.703$\pm$0.011 \\
$\bm{B}$ (full contextual information)& \textbf{0.711$\pm$0.018}\\
\hline
\end{tabular}
\label{tab:part}
\end{table}

\subsection{Case Study}\label{sec:case}
To further explore what information in the conversation sub-branch (i.e., the context) may contribute to the performance of the model, we take three representative instances with different stance labels as examples. To evaluate the contributions of words in the context, we leveraged the PyCantonese \cite{pycantonese} library for Cantonese word segmentation. Recall that $\bm{x}_{i}$ is an instance whose stance needs to be predicted. We define each Cantonese word in each instance $\bm{x}_{u}$ in the sub-branch $\bm{B}$ for $\bm{x}_{i}$  as a \textit{text span} from the $p^{\text{th}}$ token to the $q^{\text{th}}$ token in $\bm{x}_{u}$: 
$$\bm{s}_{u,pq} = [w_{u,p},\cdots,w_{u,q}],$$ 
where $1\leq u$ \textless $ i$, $p\leq q \leq n_{u}$. Recall that $n_u$ is the number of tokens in $\bm{x}_{u}$. 
Then, we mask each $\bm{s}_{u,pq}$, namely the tokens from $w_{u,p}$ to $w_{u,q}$, in each $\bm{x}_{u}$ one after another respectively. The corresponding instance with the Cantonese word $\bm{s}_{u,pq}$ masked is $$
x_{u,pq} = [\cdots, w_{u,p-1}, \text{[MASK]}, \cdots,\text{[MASK]}, w_{u,q+1},\cdots].
$$
For each $\bm{s}_{u,pq}$, we replace $\bm{x}_{u}$ in the original sub-branch $\bm{B}$ with $x_{u,pq}$ to obtain another sub-branch 
$$\bm{B}_{u,pq} = [\bm{x}_{1},\cdots,\bm{x}_{u-1}, \bm{x}_{u,pq},\bm{x}_{u+1},\cdots,\bm{x}_{i}].$$ 
Let $\hat{y}$ and $\hat{y}_{u,pq}$ be the confidence (i.e., the probability value of a predicted stance label) of $\bm{B}$ and $\bm{B}_{u,pq}$ returned by our Branch-BERT model when predicting the stance of $\bm{x}_{i}$, respectively. $\bm{s}_{u,pq}$'s contribution is thus $c_{u,pq} = \hat{y}-\hat{y}_{u,pq}$. If $c_{u,pq} \geq 20\% \cdot \hat{y}$, we say $\bm{s}_{u,pq}$ is a \emph{keyword for $\bm{x}_{i}$} \cite{sample}. We analyze the three instances with three different stances as follows. 

\textbf{Favor}. We take Comment 4 in Branch 1 in Fig.~\ref{fig:thread} as an example which has a favorable stance. 
We can observe from Fig.~\ref{fig:thread} that Comment 4 is very short, which, besides expressing the supportiveness to Comment 3, lacks valid information to infer its stance towards the target. 
Table~\ref{tab:key} illustrates the keywords for Comment 4 in Branch 1 in Fig.~\ref{fig:thread} and their corresponding contributions to the model's performance. 
First of all, the keywords for Comment 4 include names of vaccine manufacturers, ``\textit{A}'', ``\textit{B}'', the verb ``\textit{vaccinated}'', and the noun ``\textit{vaccine}'' that appear in the post and comments in the corresponding sub-branch. Such words are strongly related to the selected target ``COVID-19 vaccination'', which may explain the good performance of our model. When looking into the contents of Comments 3 and 4, the favorable stance of Comment 3 is expressed by saying that the media that ``talks bad'' about the vaccine has ulterior motives, which is acknowledged as true by Comment 4. This accounts for its predicted stance. In addition, we find that masking the words ``\textit{They have ulterior motives}'' decreases the confidence the most (by 26.74\%) among all the keywords, and that masking ``\textit{talk bad}'' reduces the confidence of the model by 15.79\%. This results indicate that the Branch-BERT model learns from not only words that strongly related to the target, but also the trigger words of stance.

\begin{table}[t]
\centering
\caption{Keywords for Comment 4 in Branch 1 in Fig.~\ref{fig:thread} and their corresponding contribution values. English translations of these words and their corresponding instances in the context are provided as well.}
\begin{tabular}{|c|c|c|c|}
\hline
\textbf{Word} & \textbf{English} & \textbf{Instance} & \textbf{Contribution} \\
\hline
\begin{CJK*}{UTF8}{gbsn}\begin{tabular}[c]{@{}c@{}}居心\\叵测\end{tabular}\end{CJK*} &  \begin{tabular}[c]{@{}c@{}}They have\\ ulterior motives\end{tabular} & Comment 3 & 26.74\% \\\hline
\begin{CJK*}{UTF8}{gbsn}疫苗\end{CJK*} &\begin{tabular}[c]{@{}c@{}} vaccine \\(the \engordnumber{3} one)\end{tabular}& Post& 19.34\% \\\hline
\begin{CJK*}{UTF8}{gbsn}大家\end{CJK*} & everyone & Comment 1& 18.64\% \\\hline
\begin{CJK*}{UTF8}{gbsn}接种\end{CJK*} & vaccinated & Post& 17.91\% \\\hline
\begin{CJK*}{UTF8}{gbsn}系统\end{CJK*} & system & Post& 17.07\% \\\hline
\begin{CJK*}{UTF8}{gbsn}品牌A\end{CJK*} &\begin{tabular}[c]{@{}c@{}} A \\(the \engordnumber{1} one)\end{tabular}& Post & 16.63\% \\\hline
\begin{CJK*}{UTF8}{gbsn}唱衰\end{CJK*} & talk bad & Comment 3& 15.79\% \\\hline
\begin{CJK*}{UTF8}{gbsn}疫苗\end{CJK*} &\begin{tabular}[c]{@{}c@{}} vaccine \\(the \engordnumber{2} one)\end{tabular}&Post & 15.61\% \\\hline
\begin{CJK*}{UTF8}{gbsn}疫苗\end{CJK*} &\begin{tabular}[c]{@{}c@{}} vaccine \\(the \engordnumber{1} one)\end{tabular}& Post & 15.42\% \\\hline
\begin{CJK*}{UTF8}{gbsn}未\end{CJK*} & never & Comment 3 & 14.82\% \\\hline
\begin{CJK*}{UTF8}{gbsn}爆出\end{CJK*} & occurred & Post & 14.59\% \\\hline
\begin{CJK*}{UTF8}{gbsn}品牌B\end{CJK*} & B & Post & 14.44\% \\\hline
\begin{CJK*}{UTF8}{gbsn}\begin{tabular}[c]{@{}c@{}}受到\\影响\end{tabular}\end{CJK*} & was affected & Post & 14.30\% \\\hline
\begin{CJK*}{UTF8}{gbsn}方向\end{CJK*} & Direction & Comment 3 & 14.02\% \\\hline
\begin{CJK*}{UTF8}{gbsn}品牌A\end{CJK*} & \begin{tabular}[c]{@{}c@{}} A \\(the \engordnumber{2} one)\end{tabular} & Post & 13.76\% \\\hline
\begin{CJK*}{UTF8}{gbsn}理解\end{CJK*} & understandable & Comment 2  & 13.41\%\\
\hline
\end{tabular}
\label{tab:key}
\end{table}

\begin{figure}[h]
    \centering
    \includegraphics[width=\linewidth]{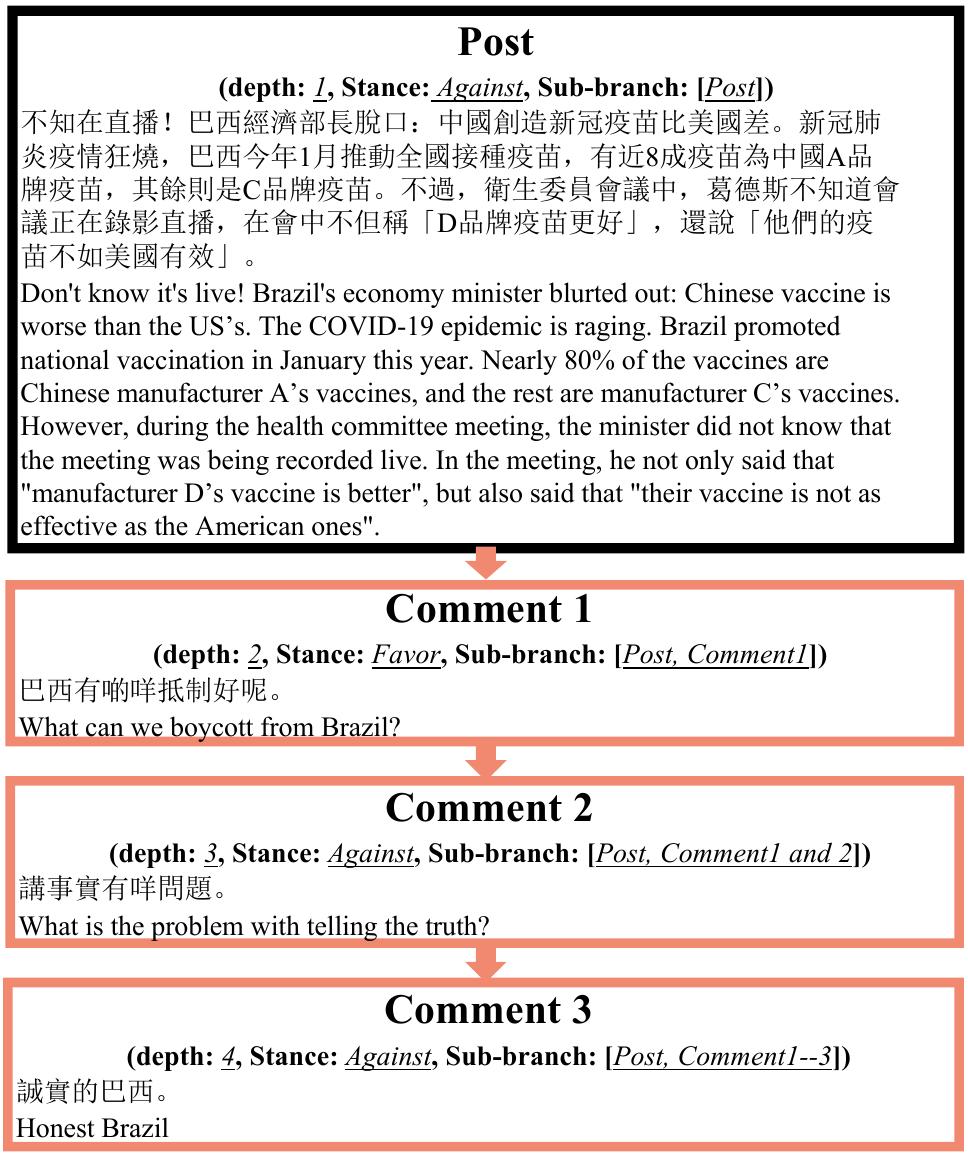}
    \caption{An example of an instance expressing an against stance and its corresponding sub-branch.}
    \label{fig:against}
\end{figure}
\begin{table}[h]
\centering
\caption{Keywords for Comment 3 in Fig.~\ref{fig:against} and their corresponding contribution values. English translations of these words and their corresponding instances in the context are provided as well.}
\begin{tabular}{|c|c|c|c|}
\hline
\textbf{Word} & \textbf{English} & \textbf{Instance} & \textbf{Contribution} \\
\hline
\begin{CJK*}{UTF8}{gbsn}疫苗\end{CJK*} &\begin{tabular}[c]{@{}c@{}} vaccine \\(the \engordnumber{6} one)\end{tabular}& Post & 21.35\% \\\hline
\begin{CJK*}{UTF8}{gbsn}有效\end{CJK*} & effective& Post & 21.33\% \\\hline
\begin{CJK*}{UTF8}{gbsn}新冠\end{CJK*} & COVID-19& Post & 21.29\% \\\hline
\begin{CJK*}{UTF8}{gbsn}直播 \end{CJK*} & live& Post & 20.27\% \\\hline
\begin{CJK*}{UTF8}{gbsn}全国\end{CJK*} & whole country & Post & 20.26\% \\\hline
\begin{CJK*}{UTF8}{gbsn}品牌A\end{CJK*} & A & Post & 20.05\% \\\hline
\begin{CJK*}{UTF8}{gbsn}不如\end{CJK*} & not as .. as & Post &  19.37\% \\\hline
\begin{CJK*}{UTF8}{gbsn}巴西\end{CJK*} &Brazil& Comment 1 & 18.23\% \\\hline
\begin{CJK*}{UTF8}{gbsn}疫苗\end{CJK*} & \begin{tabular}[c]{@{}c@{}} vaccine \\(the \engordnumber{1} one)\end{tabular}& Post& 18.19\% \\\hline
\begin{CJK*}{UTF8}{gbsn}接种\end{CJK*} & vaccination & Post& 17.78\% \\\hline
\begin{CJK*}{UTF8}{gbsn}委员会\end{CJK*} & committee & Post& 19.15\% \\\hline
\begin{CJK*}{UTF8}{gbsn}品牌 C\end{CJK*} & C & Post & 16.84\% \\\hline
\begin{CJK*}{UTF8}{gbsn}差\end{CJK*} & worse& Post& 16.39\% \\\hline
\begin{CJK*}{UTF8}{gbsn}事实\end{CJK*} & truth& Comment 2& 16.09\% \\\hline
\begin{CJK*}{UTF8}{gbsn}疫情\end{CJK*} & epidemic & Post& 15.89\% \\\hline
\begin{CJK*}{UTF8}{gbsn}部长\end{CJK*} & minister & Post& 15.47\% \\\hline
\end{tabular}
\label{tab:key2}
\end{table}

\textbf{Against}. Fig.~\ref{fig:against} illustrates a sub-branch, in which Comment 3 expresses its against stance towards the target COVID-19 vaccination. Table~\ref{tab:key2} illustrates the keywords for Comment 3 and their corresponding contributions to the model's performance. Similar to the first example, the keywords in this case contain names of vaccine manufacturers ``\textit{A}'', ``\textit{C}'' and the noun ``\textit{vaccine}'', which are strongly relevant to the target. In addition, negative words like ``\textit{worse}'' and ``\textit{not as effective as}'' are also keywords. The result further validates that Branch-BERT model learns from the words related to the target and the trigger words of stance.

\begin{figure}[h]
    \centering
    \includegraphics[width=\linewidth]{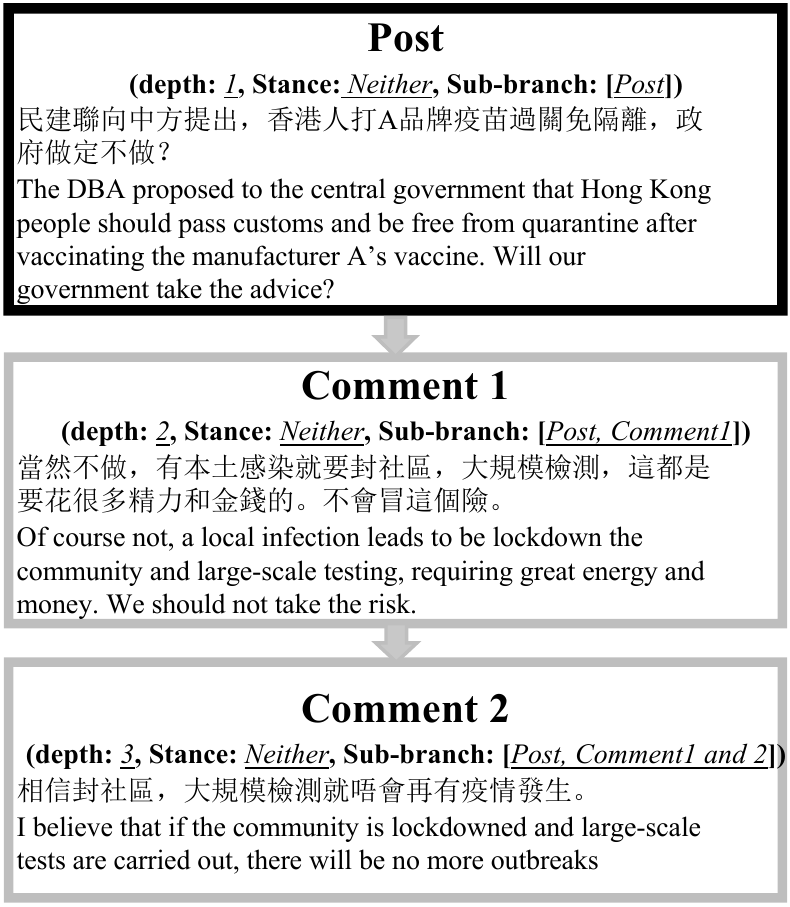}
    \caption{An example of an instance expressing neither a favorable nor against stance towards the target COVID-19 vaccination and its corresponding sub-branch.}
    \label{fig:neither}
\end{figure}

\textbf{Neither}. Fig.~\ref{fig:neither} illustrates an example expressing neither a favorable nor against stance towards the target COVID-19 vaccination and its corresponding sub-branch. Looking into the contents, we can observe that although the target is mentioned in the post, Comment 2 expresses a favorable stance towards ``\textit{lockdown community}'' and ``\textit{large-scale testing}'' instead of our target. This accounts for the neither label of Comment 2.
Table~\ref{tab:key3} illustrates the keywords for Comment 2 in Fig.~\ref{fig:neither} and their corresponding contributions to the model's performance. Same as those in the previous two cases, the names of vaccine manufacturers ``\textit{A}'' and the noun ``\textit{vaccine}'' are mentioned in the post, these words are not keywords for Comment 2. This may explain why the model assigns a predicted stance label of neither in this case. Instead, the keywords for Comment 2 include ``\textit{community}'', ``\textit{large-scale}'', and ``\textit{lockdown}''. Such words are strongly related to targets of the stance expressed in Comment 2.

\begin{table}[h]
\centering
\caption{Keywords for Comment 2 in Fig.~\ref{fig:neither} and their corresponding contribution values. English translations of these words and their corresponding instances in the context are provided as well.}
\begin{tabular}{|c|c|c|c|}
\hline
\textbf{Word} & \textbf{English} & \textbf{Instance}  & \textbf{Contribution} \\
\hline
\begin{CJK*}{UTF8}{gbsn}社区\end{CJK*} & community & Comment 1 & 23.05\% \\\hline
\begin{CJK*}{UTF8}{gbsn}感染\end{CJK*} & infection& Comment 1 & 19.26\% \\\hline
\begin{CJK*}{UTF8}{gbsn}一个\end{CJK*} & one & Comment 1 & 18.61\% \\\hline
\begin{CJK*}{UTF8}{gbsn}香港人\end{CJK*} & Hong Kong people& Post& 18.59\% \\\hline
\begin{CJK*}{UTF8}{gbsn}大規模\end{CJK*} & large-scale& Comment 1 & 16.19\% \\\hline
\begin{CJK*}{UTF8}{gbsn}封\end{CJK*} & lockdown& Comment 1 & 13.64\% \\\hline
\begin{CJK*}{UTF8}{gbsn}隔离\end{CJK*} & quarantine & Post& 12.47\% \\\hline
\begin{CJK*}{UTF8}{gbsn}金钱\end{CJK*} & money& Comment 1& 11.37\% \\\hline
\end{tabular}
\label{tab:key3}
\end{table}

\section{Conclusion}\label{sec:conclusion}
Given a practical stance detection scenario on social media platforms, we propose the conversational stance detection task which is to detect stances of the posts or comments in conversation threads on these platforms. To address this task, we constructed a Cantonese benchmarking CSD dataset with annotations of stances and conversation threads among the instances. Furthermore, we propose the Branch-BERT model that incorporates information in conversation threads in target-specific stance detection. Experimental results on CSD dataset confirm the effectiveness of using contextual information in detecting target-specific stances. Our study paves the way for future stance detection on social media platforms and sheds light on how to utilize contextual information in online conversations effectively. 
For possible future research, 
one interesting direction is to extend our proposed conversational stance detection task to 
use additional contextual information, such as comments in other branches in the conversation threads, or the behavioral information of the users who post the posts and comments. 

\bibliographystyle{IEEEtran}
\bibliography{main}

\end{document}